\documentclass{article}

\usepackage{microtype}
\usepackage{graphicx}
\usepackage{subfigure}
\usepackage{booktabs} 

\usepackage{hyperref}

\usepackage[accepted]{icml2023}

\usepackage{amsmath}
\usepackage{amssymb}
\usepackage{mathtools}
\usepackage{amsthm}

\usepackage[capitalize,noabbrev]{cleveref}

\theoremstyle{plain}
\newtheorem{theorem}{Theorem}[section]
\newtheorem{proposition}[theorem]{Proposition}

\theoremstyle{definition}

\theoremstyle{remark}

\usepackage[textsize=tiny]{todonotes}

\newcommand*\diff{\mathop{}\!\mathrm{d}}
\newcommand*\wien{\mathrm{w}}

\icmltitlerunning{Diffusion Based Representation Learning}

\begin{document}

\twocolumn[
\icmltitle{Diffusion Based Representation Learning}



\icmlsetsymbol{equal}{*}
\icmlsetsymbol{senior}{$\dagger$}

\begin{icmlauthorlist}
\icmlauthor{Sarthak Mittal}{equal,mila,udem}
\icmlauthor{Korbinian Abstreiter}{equal,eth}
\icmlauthor{Stefan Bauer}{helmholtz,tum}
\icmlauthor{Bernhard Schölkopf}{mpi}
\icmlauthor{Arash Mehrjou}{senior,eth,mpi}
\end{icmlauthorlist}

\icmlaffiliation{mila}{Mila}
\icmlaffiliation{udem}{Université de Montréal}
\icmlaffiliation{eth}{ETH Zürich}
\icmlaffiliation{helmholtz}{Helmholtz AI}
\icmlaffiliation{tum}{Technical University of Munich}
\icmlaffiliation{mpi}{Max Planck Institute for Intelligent Systems}

\icmlcorrespondingauthor{Sarthak Mittal}{sarthmit@gmail.com}
\icmlcorrespondingauthor{Arash Mehrjou}{arash@distantvantagepoint.com}

\icmlkeywords{Machine Learning, ICML, Diffusion Models, Representation Learning}

\vskip 0.3in
]




\printAffiliationsAndNotice{$^*$Equal contribution, $^\dagger$ Senior authorship, } 

\begin{abstract}
    Diffusion-based methods, represented as stochastic differential equations on a continuous-time domain, have recently proven successful as non-adversarial generative models.
    Training such models relies on denoising score matching, which can be seen as multi-scale denoising autoencoders.
    Here, we augment the denoising score matching framework to enable representation learning without any supervised signal.
    GANs and VAEs learn representations by directly transforming latent codes to data samples.
    In contrast, the introduced diffusion-based representation learning relies on a new formulation of the denoising score matching objective and thus encodes the information needed for denoising.
    We illustrate how this difference allows for manual control of the level of details encoded in the representation.
    Using the same approach, we propose to learn an infinite-dimensional latent code that achieves improvements on state-of-the-art models on semi-supervised image classification. We also compare the quality of learned representations of diffusion score matching with other methods like autoencoder and contrastively trained systems through their performances on downstream tasks. Finally, we also ablate with a different SDE formulation for diffusion models and show that the benefits on downstream tasks are still present on changing the underlying differential equation.
\end{abstract}

\section{Introduction}
Diffusion-based models have recently proven successful for generating 
images \citep{sohldickstein2015deep, DBLP:journals/corr/abs-2006-09011, song2020denoising}, graphs \citep{niu2020permutation}, shapes \citep{cai2020learning}, and audio \citep{chen2020wavegrad, kong2021diffwave}.
Two promising approaches apply step-wise perturbations to samples of the data distribution until the perturbed distribution matches a known prior \citep{DBLP:journals/corr/abs-1907-05600, DBLP:journals/corr/abs-2006-11239}. A model is then trained to estimate the reverse process, which transforms samples of the prior to samples of the data distribution \citep{saremi2018deep}. Diffusion models were further refined \citep{DBLP:journals/corr/abs-2102-09672, luhman2021knowledge} and even achieved better image sample quality than GANs \citep{dhariwal2021diffusion, ho2021cascaded, mehrjou2017annealed}.
Further, \citeauthor{song2021scorebased} showed that these frameworks are discrete versions of continuous-time perturbations modeled by stochastic differential equations and proposed a diffusion-based generative modeling framework on continuous time. Unlike generative models such as GANs and various forms of autoencoders, the original form of diffusion models does not come with a fixed architectural module that captures the representations of the data samples.

Learning desirable representations has been an integral component of generative models such as GANs and VAEs \citep{bengio2013representation, radford2016unsupervised, chen2016infogan, DBLP:journals/corr/abs-1711-00937, donahue2019large, pmlr-v119-chen20s, scholkopf2021toward}.
Recent works on visual representation learning achieve impressive performance on the downstream task of classification by applying contrastive learning \citep{DBLP:journals/corr/abs-2006-10029, grill2020bootstrap, chen2020exploring, caron2021unsupervised,chen2020simple}. However, contrastive learning requires additional supervision of augmentations that preserve the content of the data, and hence these approaches are not directly comparable to representations learned through generative systems like Variational Autoencoders~\citep{kingma2013auto,rezende2014stochastic} and the current work which are considered \emph{fully} unsupervised. Moreover, training the encoder to output similar representation for different views of the same image removes information about the applied augmentations, thus the performance benefits are limited to downstream tasks that do not depend on the augmentation, which has to be known beforehand. Hence our proposed algorithm does not restrict the learned representations to specific downstream tasks and solves a more general problem instead. We provide a summary of contrastive learning approaches in Appendix \ref{sec:contrastive_learning}.
Similar to our approach, Denoising Autoencoders (DAE) \citep{10.1145/1390156.1390294} can be used to encode representations that can be manually controlled by adjusting the noise scale \citep{geras2015scheduled, inproceedingsDAE, 10.1007/s11704-016-6107-0}. Note that, unlike DAEs, the encoder in our approach does not receive noisy data as input, but instead extracts features based on the clean images. For example, this key difference allows DRL to be used to limit the encoding to fine-grained features when focusing on low noise levels, which is not possible with DAEs.

Recently, there have been some works that rely on additional encoders in the model architecture of diffusion based models~\citep{preechakul2022diffusion,mittal2021symbolic,sinha2021}. \citet{sinha2021} considers an autoencoder based setup with the diffusion model defining the prior whereas \citet{pandey2022diffusevae} considers the opposite where a diffusion model is used to further improve the decoded samples from a VAE. \citet{preechakul2022diffusion} is a concurrent work that is closest to our setup, however, instead of relying on time-conditioned encoder, they rely only on an unconditional encoder. Further, they concentrate more on generation-based tasks while our approach focuses more on evaluating the representations learned for downstream tasks.

\begin{figure}[t!]
\centering
\includegraphics[width=0.6\columnwidth]{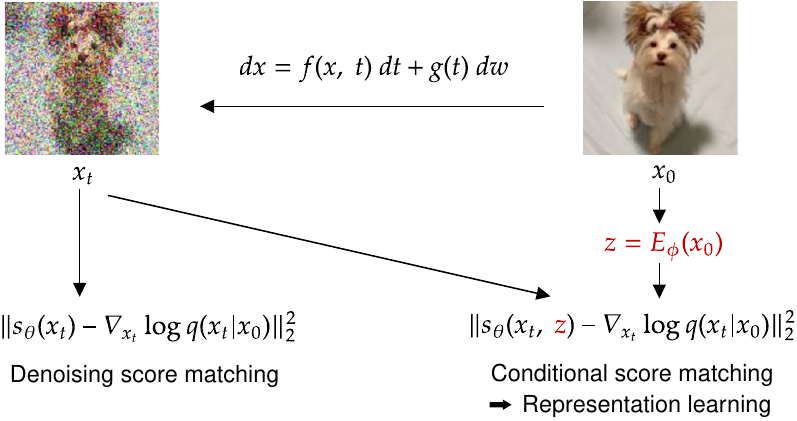}
\caption{Conditional score matching with a parametrized latent code is representation learning. Denoising score matching estimates the score at each $x_t$; we add a latent representation $z$ of the clean data $x_0$ as additional input to the score estimator.}
\label{fig:schem_diag}
\vspace{-4mm}
\end{figure}
The main contributions of this work are
\begin{itemize}
    \item We present an alternative formulation of the denoising score matching objective, showing that the objective cannot be reduced to zero. We leverage this property to learn representations for downstream tasks.
    \item We introduce Diffusion-based Representation Learning (DRL), a novel framework for representation learning in diffusion-based generative models. We show how this framework allows for manual control of the level of details encoded in the representation through an infinite-dimensional code. We evaluate the proposed approach on downstream tasks using the learned representations directly as well as using it as a pre-training step for semi-supervised image classification, thereby improving state-of-the-art approaches for the latter.
    \item We evaluate the effect of the initial noise scale and achieve significant improvements in sampling speed, which is a bottleneck in diffusion-based generative models compared with GANs and VAEs, without sacrificing image quality.
\end{itemize}

The implementation of our method and the code to reproduce the experimental results are publicly available in \url{https://github.com/amehrjou/Diffusion-Based-Representation-Learning}.

\begin{figure*}[t!]
\centering
\begin{minipage}{\columnwidth}
\subfigure{
\makebox[0.24\columnwidth][c]{\includegraphics[width=0.24\columnwidth]{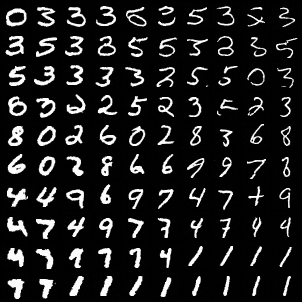}}
} 
\hspace{1mm}
\subfigure{\makebox[0.2\columnwidth][c]{\includegraphics[width=0.2\columnwidth]{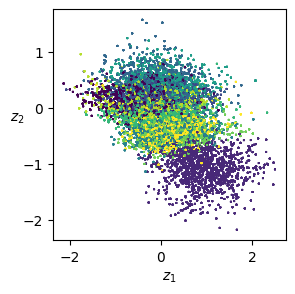}}}
\hspace{2mm}
\subfigure{\makebox[0.24\columnwidth][c]{\includegraphics[width=0.24\columnwidth]{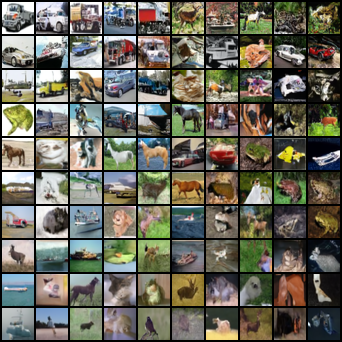}}}
\hspace{1mm}
\subfigure{\makebox[0.2\columnwidth][c]{\includegraphics[width=0.2\columnwidth]{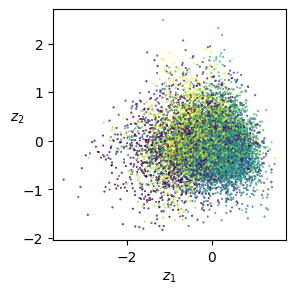}}}
\end{minipage}
\begin{minipage}{\columnwidth}
\subfigure{
\makebox[0.24\columnwidth][c]{\includegraphics[width=0.24\columnwidth]{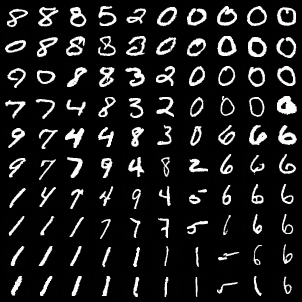}}
} 
\hspace{1mm}
\subfigure{\makebox[0.2\columnwidth][c]{\includegraphics[width=0.2\columnwidth]{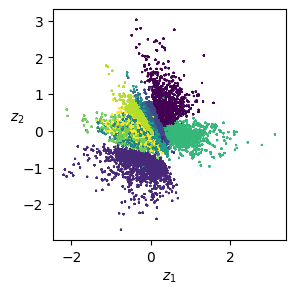}}}
\hspace{2mm}
\subfigure{\makebox[0.24\columnwidth][c]{\includegraphics[width=0.24\columnwidth]{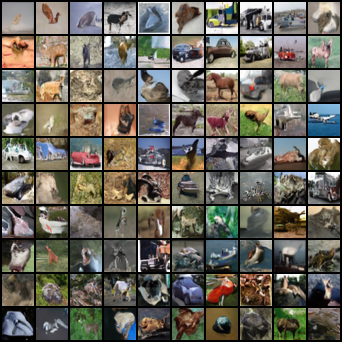}}}
\hspace{1mm}
\subfigure{\makebox[0.2\columnwidth][c]{\includegraphics[width=0.2\columnwidth]{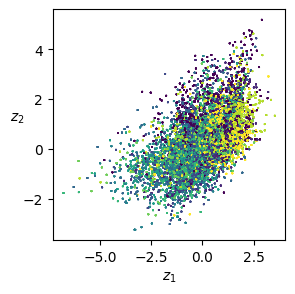}}}
\end{minipage} 
\caption{Results of proposed DRL models trained on MNIST and CIFAR-10 with point clouds visualizing the latent representation of test samples, colored according to the digit class. The models are trained with \textbf{Left:} uniform sampling of $t$ and \textbf{Right:} a focus on high noise levels. Samples are generated from a grid of latent values ranging from -1 to 1.}
\label{fig:vis_drl}
\vspace{-4mm}
\end{figure*}
\subsection{Diffusion-based generative modeling}
\label{sec:diff_process}
We first give a brief overview of the technical background for the framework of the diffusion-based generative model as described in \cite{song2021scorebased}.
The forward diffusion process of the data is modeled as an SDE on a continuous-time domain $t \in [0, T]$. Let $x_0 \in \mathbb{R}^d$ denote a sample from the data distribution $x_0 \sim p_0$, where $d$ is the data dimension. The trajectory $(x_t)_{t\in[0, T]}$ of data samples is a function of time determined by the diffusion process. The SDE is chosen such that the distribution $p_{0T}(x_T | x_0)$ for any sample $x_0 \sim p_0$ can be approximated by a known prior distribution. Notice that the subscript $0T$ of $p_{0T}$ refers to the conditional distribution of the diffused data at time $T$ given the data at time $0$.
For simplicity we limit the remainder of this paper to the so-called Variance Exploding SDE \citep{song2021scorebased}, that is,
\begin{align}
\label{eq:forward_sde}
\diff x = f(x, t)\diff t + g(t) \diff \wien := \sqrt{\frac{\diff [\sigma^2(t)]}{\diff t}} \diff \wien,
\end{align}
where $\wien$ is the standard Wiener process.
The perturbation kernel of this diffusion process has a closed-form solution being 
$p_{0t}(x_t | x_0) = \mathcal{N}(x_t; x_0, [\sigma^2(t) - \sigma^2(0)]I)$.
It was shown by \citet{ANDERSON1982313} that the reverse diffusion process is the solution to the following SDE:
\begin{align}
\diff x = [f(x, t) - g^2(t) \nabla_x \log p_t(x)] \diff t + g(t) \diff \overline{\wien},
\end{align}
where $\overline{\wien}$ is the standard Wiener process when the time moves backwards. Thus, given the score function $\nabla_x \log p_t(x)$ for all $t \in [0, T]$, we can generate samples from the data distribution $p_0(x)$.
In order to learn the score function, the simplest objective is Explicit Score Matching (ESM) \citep{hyvarinen2005estimation}, that is,
\begin{align}
\label{eq:esm}
    \mathbf{E}_{x_t} \left[ \lVert s_\theta(x_t, t) - \nabla_{x_t} \log p_{t}(x_t) \rVert^2_2 \right].
\end{align}
Since the ground-truth score function $\nabla_{x_t} \log p_{t}(x_t)$ is generally not known, one can apply denoising score matching (DSM) \citep{6795935}, which is defined as the following:
\begin{align}
\begin{aligned}
\label{eq:dsm}
J^{DSM}_t(\theta) = & \mathbf{E}_{x_0} \large\{ \mathbf{E}_{x_t | x_0} \large[
    \lVert s_\theta(x_t, t) \\&- \nabla_{x_t} \log p_{0t}(x_t | x_0) \rVert^2_2 \,\large] \large\}.
\end{aligned}
\end{align}
The training objective over all $t$ is augmented by \citet{song2021scorebased} with a time-dependent positive weighting function $\lambda(t)$, that is, 
$J^{DSM}(\theta) = \mathbf{E}_{t} \left[ \lambda(t) J^{DSM}_t(\theta) \right]$.
One can also achieve class-conditional generation in diffusion-based models by training an additional time-dependent classifier $p_t(y|x_t)$ \citep{song2021scorebased}). 
In particular, the conditional score for a fixed $y$ can be expressed as the sum of the unconditional score and the score of the classifier, that is, 
$\nabla_{x_t} \log p_t(x_t|y) = \nabla_{x_t} \log p_t(x_t) +\nabla_{x_t} \log p_t(y|x_t)$.
We take motivation from an alternative way to allow for controllable generation, which, given supervised samples $(x, y(x))$, uses the following training objective for each time $t$
\begin{align}
\begin{aligned}
\label{eq:dsm_cond}
J^{CSM}_t(\theta) = \mathbf{E}_{x_0} \large\{ &\mathbf{E}_{x_t | x_0} \large[
    \lVert s_\theta(x_t, t, y(x_0)) \\&-\nabla_{x_t} \log p_{0t}(x_t | x_0) \rVert^2_2 \,\large] \large\}.
\end{aligned}
\end{align}
The objective in Equation \ref{eq:dsm_cond} is minimized if and only if the model equals the conditional score function $\nabla_{x_t} \log p_{t}(x_t | y(x_0) = \hat{y})$ for all labels $\hat{y}$. 
\section{Diffusion-based Representation Learning}
\label{sec:reprlearning}
We begin this section by presenting an alternative formulation of the Denoising Score Matching (DSM) objective, which shows that this objective cannot be made arbitrarily small. Formally, the formula of the DSM objective can be rearranged as 
\begin{align}
\begin{aligned}
\label{eq:dsmesm}
J^{DSM}_t(\theta) &= \mathbf{E}_{x_0} \{ \mathbf{E}_{x_t|x_0}\big[\lVert s_\theta(x_t, t) - \nabla_{x_t} \log p_t(x_t) \rVert^2_2 \\&+
\lVert \nabla_{x_t} \log p_{0t}(x_t | x_0) - \nabla_{x_t} \log p_t(x_t) \rVert^2_2 \big] \}.
\end{aligned}
\end{align}
The above formulation holds, because the DSM objective in Equation \ref{eq:dsm} is minimized when $\forall x_t: s_\theta(x_t, t) = \nabla_{x_t} \log p_t(x_t)$, and differs from ESM in Equation \ref{eq:esm} only by a constant \citep{6795935}. Hence, the constant is equal to the minimum achievable value of the DSM objective.
A detailed proof is included in the Appendix \ref{proof:dsm_formulation}.

It is noteworthy that the second term in the right-hand side of the Equation \ref{eq:dsmesm} does not depend on the learned score function of $x_t$ for every $t\in[0, T]$. Rather, it is influenced by the diffusion process that generates $x_t$ from $x_0$.
This observation has not been emphasized previously, probably because it has no direct effect on the learning of the score function, which is handled by the second term in the Equation \ref{eq:dsmesm}. However, the additional constant has major implications for finding other hyperparameters such as the function $\lambda(t)$ and the choice of $\sigma(t)$ in the forward SDE. As \cite{kingma2021variational} shows, changing the integration variable from time to signal-to-noise ratio (SNR) simplifies the diffusion loss such that it only depends on the end values of SNR. Hence, the loss is invariant to the intermediate values of the noise schedule. However, the weight functions $\lambda(\cdot)$ is still an important hyper-parameter whose choice might be affected by the non-vanishing constant in Equation \ref{eq:dsmesm}.

To the best of our knowledge, there is no known theoretical justification for the values of $\sigma(t)$.
While these hyperparameters could be optimized in ESM using gradient-based learning, this ability is severely limited by the non-vanishing constant in Equation \ref{eq:dsmesm}.

Even though the non-vanishing constant in the denoising score matching objective presents a burden in multiple ways such as hyperparameter search and model evaluation, it provides an opportunity for latent representation learning, which will be described in the following sections. We note that this is different from \citet{sinha2021,mittal2021diffusion} as they consider a Variational Autoencoder model followed by diffusion in the latent space, where their representation learning objective is still guided by reconstruction. Contrary to this, our representation learning approach does not utilize a variational autoencoder model and is guided by denoising instead. Our approach is similar to \citet{preechakul2022diffusion} but we also condition the encoder system on the time-step, thereby improving representation capacity and leading to parameterized curve-based representations.
\begin{figure*}[t!]
\centering
\begin{minipage}{\columnwidth}
\subfigure{
\makebox[0.24\columnwidth][c]{\includegraphics[width=0.24\columnwidth]{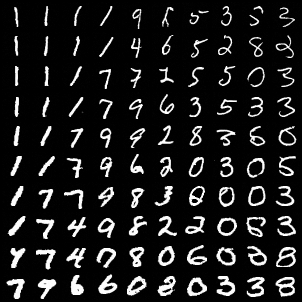}}
} 
\hspace{1mm}
\subfigure{\makebox[0.2\columnwidth][c]{\includegraphics[width=0.2\columnwidth]{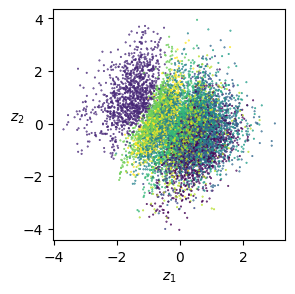}}}
\hspace{2mm}
\subfigure{\makebox[0.24\columnwidth][c]{\includegraphics[width=0.24\columnwidth]{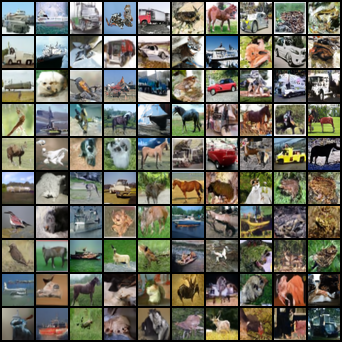}}}
\hspace{1mm}
\subfigure{\makebox[0.2\columnwidth][c]{\includegraphics[width=0.2\columnwidth]{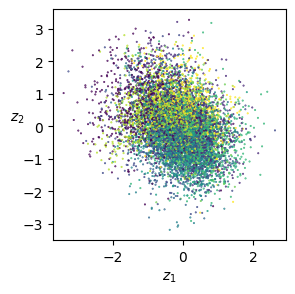}}}
\end{minipage} 
\begin{minipage}{\columnwidth}
\subfigure{
\makebox[0.24\columnwidth][c]{\includegraphics[width=0.24\columnwidth]{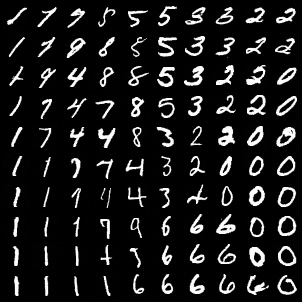}}
} 
\hspace{1mm}
\subfigure{\makebox[0.2\columnwidth][c]{\includegraphics[width=0.2\columnwidth]{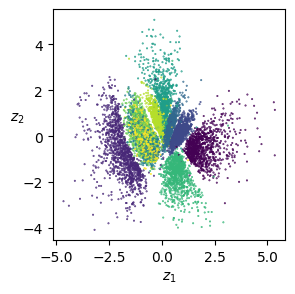}}}
\hspace{2mm}
\subfigure{\makebox[0.24\columnwidth][c]{\includegraphics[width=0.24\columnwidth]{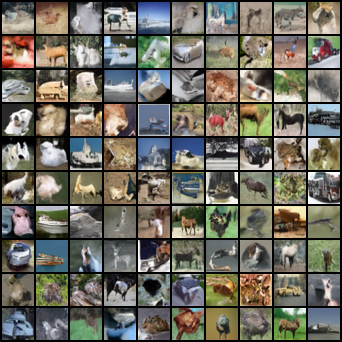}}}
\hspace{1mm}
\subfigure{\makebox[0.2\columnwidth][c]{\includegraphics[width=0.2\columnwidth]{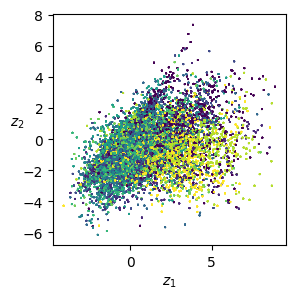}}}
\end{minipage} 
\caption{Results of proposed VDRL models trained on MNIST and CIFAR-10 with point clouds visualizing the latent representation of test samples, colored according to the digit class. The models are trained with \textbf{Left:} uniform sampling of $t$ and \textbf{Right:} a focus on high noise levels. Samples are generated from a grid of latent values ranging from -2 to 2.}
\label{fig:vis_vdrl}
\vspace{-4mm}
\end{figure*}
\subsection{Learning latent representations}
Since supervised data is limited and rarely available, we propose to learn a labeling function $y(x_0)$ at the same time as optimizing the conditional score matching objective in Equation \ref{eq:dsm_cond}. In particular, we represent the labeling function as a trainable encoder $E_\phi: \mathbb{R}^d\to \mathbb{R}^{c}$, where $E_\phi(x_0)$ maps the data sample $x_0$ to its corresponding code in the $c$-dimensional latent space. The code is then used as additional input to the score model. Formally, the proposed learning objective for Diffusion-based Representation Learning (DRL) is the following:
\begin{align}
\begin{aligned}
\label{eq:repr_obj}
J^{DRL}(\theta, \phi) &= \mathbf{E}_{t, x_0, x_t} \large[\lambda(t)
    \lVert s_\theta(x_t, t, E_\phi(x_0)) \\&- \nabla_{x_t} \log p_{0t}(x_t | x_0) \rVert^2_2 \, + \gamma\lVert E_\phi(x_0)\rVert_1\large]
\end{aligned}
\end{align}
where we add a small amount of $L_1$ regularization, controlled by $\gamma$, on the output of the trainable encoder.

To get a better idea of the above objective, we provide an intuition for the role of $E_\phi(x_0)$ in the input of the model. The model $s_\theta(\cdot, \cdot, \cdot): \mathbb{R}^d\times \mathbb{R}\times \mathbb{R}^c\to \mathbb{R}^d$ is a vector-valued function whose output points to different directions based on the value of its third argument. In fact, $E_\phi(x_0)$ selects the direction that best recovers $x_0$ from $x_t$. Hence, when optimizing over $\phi$, the encoder learns to extract the information from $x_0$ in a reduced-dimensional space that helps recover $x_0$ by denoising $x_t$. 

We show in the following that Equation \ref{eq:repr_obj} is a valid representation learning objective.
The score of the perturbation kernel $\nabla_{x_t} \log p_{0t}(x_t | x_0)$ is a function of only $t$, $x_t$ and $x_0$. 
Thus, the objective can be reduced to zero if all information about $x_0$ is contained in the latent representation $E_\phi(x_0)$. 
When $E_\phi(x_0)$ has no mutual information with $x_0$, the objective can only be reduced up to the constant in Equation \ref{eq:dsmesm}. Hence, our proposed formulation takes advantage of the non-zero lower-bound of Equation \ref{eq:dsmesm}, which can only vanish when the encoder $E_\phi(\cdot)$ properly distills information from the unperturbed data into a latent code, which is an additional input to the score model. 
These properties show that Equation \ref{eq:repr_obj} is a valid objective for representation learning.

\label{sec:control}
Our proposed representation learning objective enjoys the continuous nature of SDEs, a property that is not available in many previous representation learning methods \citep{radford2016unsupervised, chen2016infogan, locatello2019challenging}. In DRL, the encoder is trained to represent the information needed to denoise $x_0$ for different levels of noise $\sigma(t)$. We hypothesize that by adjusting the weighting function $\lambda(t)$, we can manually control the granularity of the features encoded in the representation and provide empirical evidence as support. 
Note that $t\to T$ is associated with higher levels of noise and the mutual information of $x_t$ and $x_0$ starts to vanish. In this case, denoising requires all information about $x_0$ to be contained in the code. In contrast, $t\to 0$ corresponds to low noise levels and hence $x_t$ contains coarse-grained features of $x_0$ and only fine-grained properties may have been washed out. Hence, the encoded representation learns to keep the information needed to recover these fine-grained details. We provide empirical evidence to support this hypothesis in Section \ref{sec:results}.

\begin{figure*}
\includegraphics[width=\textwidth,trim={0 0 0 0},clip]{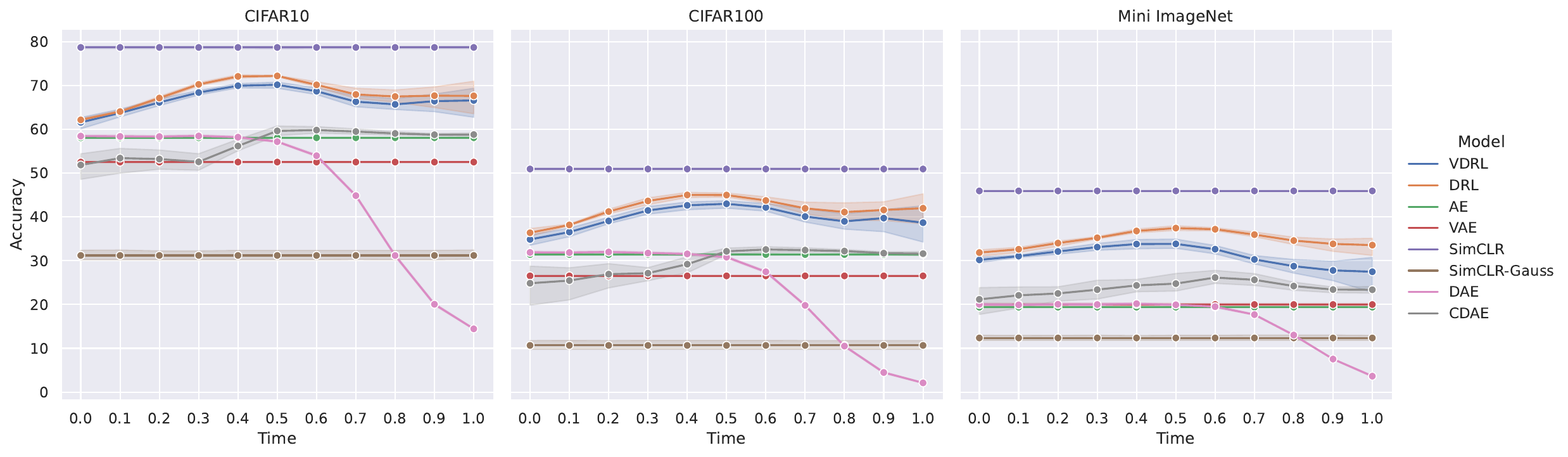}
\vspace{-7mm}
\caption{Comparing the performance of the proposed diffusion-based representations (DRL and VDRL) with the baselines that include autoencoder (AE), variational autoencoder (VAE), simple contrastive learning (simCLR) and its restricted variant (simCLR-Gauss) which exclude domain-specific data augmentation from the original simCLR algorithm.}
\label{fig:ve}
\vspace{-4mm}
\end{figure*}
It is noteworthy that $E_\phi$ does not need to be a deterministic function and can be a probabilistic map similar to the encoder of VAEs. In principle, it can be viewed as an information channel that controls the amount of information that the diffusion model receives from the initial point of the diffusion process. With this perspective, any deterministic or stochastic function that can manipulate $I(x_t, x_0)$, the mutual information between $x_0$ and $x_t$, can be used. This opens up the room for stochastic encoders similar to VAEs which we call Variational Diffusion-based Representation Learning (VDRL). The formal objective of VDRL is
\begin{align}
\label{eq:repr_obj_vdrl}
J^{VDRL}(\theta, \phi) &= \mathbf{E}_{t, x_0, x_t} \large[ \mathbf{E}_{z \sim E_\phi(Z|x_0)} \large[\lambda(t)
    \lVert s_\theta(x_t, t, z) \notag\\&\quad- \nabla_{x_t} \log p_{0t}(x_t | x_0) \rVert^2_2 \,\large] \\&
    \quad + \mathcal{D}_{\rm KL}(E_\phi(Z|x_0) || \mathcal{N}(Z;0,I) \large] \notag
\end{align}

\subsection{Infinite-dimensional representation of data}
\label{sec:time_repr}
We now present an alternative version of DRL where the representation is a function of time. Instead of emphasizing on different noise levels by weighting the training objective, as done in the previous section, we can provide the time $t$ as input to the encoder. Formally, the new objective is
\begin{align}
    \mathbf{E}_{t, x_0, x_t} &\large[\lambda(t)
    \lVert s_\theta(x_t, t, E_\phi(x_0, t)) \notag\\&- \nabla_{x_t} \log p_{0t}(x_t | x_0) \rVert^2_2 \, + \gamma \lVert E_\phi(x_0, t)\rVert_1\large]
\label{eq:time_repr_obj}
\end{align}
where $E_\phi(x_0)$ in Equation \ref{eq:repr_obj} is replaced by $E_\phi(x_0, t)$. Intuitively, it allows the encoder to extract the necessary information of $x_0$ required to denoise $x_t$ for any noise level. This leads to richer representation learning since normally in autoencoders or other \emph{static} representation learning methods, the input data $x_0\in \mathbb{R}^d$ is mapped to a single point $z\in \mathbb{R}^c$ in the latent space. In contrast, we propose a richer representation where the input $x_0$ is mapped to a curve in $\mathbb{R}^c$ instead of a single point. Hence, the learned latent code is produced by the map $x_0\to (E_\phi(x_0, t))_{t\in[0, T]}$ where the infinite-dimensional object $(E_\phi(x_0, t))_{t\in[0, T]}$ is the encoding for $x_0$. 

\begin{proposition}
\label{theo:infinite_repr}
For any downstream task, the infinite-dimensional code $(E_\phi(x_0, t))_{t\in[0, T]}$ learned using the objective in Equation \ref{eq:time_repr_obj} is at least as good as finite-dimensional static codes learned by the reconstruction of $x_0$.
\end{proposition}
\textit{Proof sketch.} Let $L_D(z, y)$ be the per-sample loss for a supervised learning task calculated for the pair $(z, y)$ where $z = z(x, t)$ is the representation learned for the input $x$ at time $t$ and $y$ is the label. The representation function is also a function of the scalar $t$ that takes values from a closed subset $U$ of $R$. For any value  $s\in U$, it is obvious that
\begin{align}
    min_{t\in U} L_D(z(x, t), y) < L_D(z(x, s), y).
\end{align}
Taking into account the extra argument $t$, the representation function $z(x, t)$ can be seen as an infinite dimensional representation. The argument $t$ actually controls which representation of $x$ has to be passed to the downstream task. The conventional representation learning algorithms correspond to choosing the $t$ argument apriori and keep it fixed independent of $x$. Here, by minimizing over $t$, the passed representation cannot be worse than the results of conventional representation learning methods. Note that $L_D(\cdot, \cdot)$ here can be any metric that we require, however gradient-based learning and optimization issues can still affect the actual performance achieved  .

The score matching objective can be seen as a reconstruction objective of $x_0$ conditioned on $x_t$. The terminal time $T$ is chosen large enough so that $x_T$ is independent of $x_0$, hence the objective for $t=T$ is equal to a reconstruction objective without conditioning. Therefore, there exists a $t\in[0,T]$ where the learned representation $E_\phi(x_0, t)$ is the same representation learned by the reconstruction objective of a vanilla autoencoder.
The full proof for Proposition \ref{theo:infinite_repr} can be found in the Appendix  \ref{theo:infinite_repr_proof} 

A downstream task can leverage this rich encoding in various ways, including the use of either the static code for a fixed $t$, or the use of the whole trajectory $(E_\phi(x_0, t))_{t\in[0, T]}$ as input.
We posit the conjecture that the proposed rich representation is helpful for downstream tasks when used for pretraining, where the value of $t$ could either be a model selection parameter or be jointly optimized with other parameters during training. We leave investigations along these directions as important future work. We show the performance of the proposed model on downstream tasks in Section \ref{sec:downstreamsimpletask}
and also evaluate it on semi-supervised image classification in Section \ref{sec:downstreamtask}.

\section{Results}
\label{sec:results}
For all experiments, we use the same function $\sigma(t), t \in [0, 1]$ as in \citet{song2021scorebased}, which is $\sigma(t) = \sigma_{\min}\left(\sigma_{\max}/\sigma_{\min}\right)^t$, where $\sigma_{\min}=0.01$ and $\sigma_{\max}=50$. Further, we use a $2$d latent space for all qualitative experiments (Section \ref{sec:qual}) and $128$ dimensional latent space for the downstream tasks (Section \ref{sec:downstreamsimpletask}) and semi-supervised image classification (Section \ref{sec:downstreamtask}). We also set $\lambda(t) = \sigma^2(t)$, which has been shown to yield the KL-Divergence objective \citep{durkan2021maximum}.
Our goal is not to produce state-of-the-art image quality, rather showcase the representation learning method. Because of that and also limited computational resources, we did not carry out an extensive hyperparameter sweep (check Appendix \ref{hyperparams} for details). Note that all experiments were conducted on a single RTX8000 GPU, taking up to 30 hours of wall-clock time, which only amounts to 15\% of the iterations proposed in \cite{song2021scorebased}.
\begin{figure*}
\includegraphics[width=\textwidth,trim={0 0 0 0},clip]{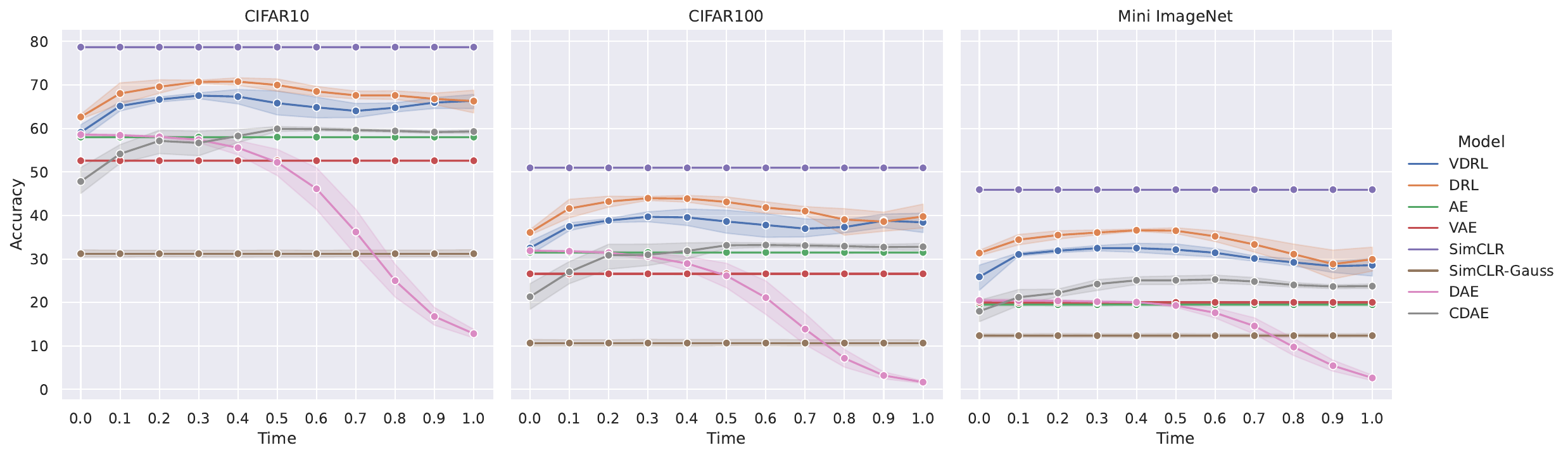}
\vspace{-7mm}
\caption{Comparing the performance of the proposed diffusion-based representations (DRL and VDRL) with the baselines that include autoencoder (AE), variational autoencoder (VAE), simple contrastive learning (simCLR) and its restricted variant (simCLR-Gauss) which exclude domain-specific data augmentation from the original simCLR algorithm.}
\label{fig:vp}
\vspace{-4mm}
\end{figure*}
\subsection{Downstream Classification}
\label{sec:downstreamsimpletask}
We directly evaluate the representations learned by different algorithms on downstream classification tasks for CIFAR10, CIFAR100, and Mini-ImageNet datasets. The representation is first learned using the proposed diffusion-based method. Then, the encoder (either deterministic or probabilistic) is frozen and a single-layered neural network is trained on top of it for the downstream prediction task. For the baselines, we consider an Autoencoder (AE), a Variational Autoencoder (VAE), two versions of Denoising Autoencoders (DAE and CDAE) and two verisons of Contrastive Learning (SimCLR\cite{chen2020simple} and SimCLR-Gauss explained below) setup to compare with the proposed methods (DRL and VDRL). Figure \ref{fig:ve} shows that DRL and VDRL outperforms autoencoder-styled baselines as well as the restricted contrastive learning baseline.

\textit{Standard Autoencoders---} Standard autoencoders (AE and VAE) rely on learning of representations of the input data using an encoder in such a way that it can be reconstructed back, using a decoder, solely based on the representation learned. Such systems can be trained without any regularization on the representation space (AE), or in a probabilistic fashion which relies on variational inference and ultimately leads to a KL-Divergence based regularization on the representation space (VAE). Figure~\ref{fig:ve} shows that the time-axis is not meaningful for such training, as expected.

\textit{Denoising Autoencoders---} While the problem of reconstruction is easily solved given a big enough network (i.e. capable of learning the identity mapping), this problem can be made harder by considering a noisy version of the data as input with the task of predicting its denoised version, as opposed to vanilla reconstruction in standard autoencoders. Such approaches are referred to as Denoising Autoencoders, and we consider its two variants. In the first variant, DAE, a noisy version of the image is given as input $x_t$ (higher $t$ implying more noise) and the task of the model is to predict the denoised version $x_0$. Since larger $t$ implies learning of representations from more noise, we can see a sharp decline in performance of DAE systems with increasing $t$ in Figure~\ref{fig:ve}. The second variant, CDAE, considers $x_t$ as the noisy input again, but predicts the denoised version based on a representation of $x_t$ combined with a learned time-conditioned representation of the true input $E_\phi(x_0, t)$, similar to the DRL setups. This approach is arguably similar to DRL with the sole difference being that $E_\phi(\cdot, \cdot)$ in DRL had the incentive of predicting the right score function, whereas in CDAE the incentive is to denoise in a single step. As highlighted in Figure~\ref{fig:ve}, the performance increases with increasing $t$ because the encoder $E_\phi(\cdot, \cdot)$ is useless in low-noise settings (as all the data is already there in the input) but becomes increasingly meaningful as noise increases.

\begin{table*}
\centering
\small
\setlength{\tabcolsep}{15pt}
\begin{tabular}{@{}lrccccccc@{}}
\toprule
&& \multicolumn{2}{c}{LaplaceNet} & \multicolumn{3}{c}{Ours}
\\
\multicolumn{2}{@{}l}{Pretraining} & 
\multicolumn{2}{c}{None} & \multicolumn{2}{c}{DRL} & VDRL 
\\
\multicolumn{2}{@{}l}{Mixup} & No & Yes & No & Yes & No
\\
\midrule
Dataset & \#labels & 
\\
\midrule
CIFAR-10 & 100 & 73.68 & 75.29 & 74.31 & 64.67 & \textbf{81.63} \\
& 500 & 91.31 & 92.53 & \textbf{92.70} & 92.31 & \textbf{92.79}\\
& 1000 & 92.59 & 93.13 & \textbf{93.24} & \textbf{93.42} & \textbf{93.60}\\
& 2000 & 94.00 & 93.96 & \textbf{94.18} & 93.91 & 93.96 \\
& 4000 & 94.73 & 94.97 & 94.75 & \textbf{95.22} & \textbf{95.00}\\
\midrule
CIFAR-100 & 1000 & 55.58 & 55.24 & \textbf{55.85} & \textbf{55.74} & \textbf{56.47}\\
& 4000 & 67.07 & 67.25 & 67.22 & \textbf{67.47} & \textbf{67.54}\\
& 10000 & 73.19 & 72.84 & \textbf{73.31} & \textbf{73.66} & \textbf{73.50}\\
& 20000 & 75.80 & 76.07 & \textbf{76.46} & \textbf{76.88} & \textbf{76.64}\\
\midrule
Mini ImageNet & 4000 & 58.40 & 58.84 & \textbf{58.95} & \textbf{59.29} & \textbf{59.14}\\
& 10000 & 66.65 & 66.80 & \textbf{67.31} & 66.63 & \textbf{67.46}\\
\bottomrule
\end{tabular}
\caption{Comparison of classifier accuracy in \% for different pretraining settings. Scores better than the SOTA model (LaplaceNet) are in \textbf{bold}. ``DRL" pretraining is our proposed representation learning, and ``VDRL" the respective version which uses a probabilistic encoder.}
\label{tab:clf_by_nlabels}
\vspace{-4mm}
\end{table*}
\textit{Restricted SimCLR---} While we compare against the standard SimCLR model, to obtain a fair comparison, we restricted the transformations used by the simCLR method to the additive pixel-wise Gaussian noise (SimCLR-Gauss) as this was the only domain-agnostic transformation in the SimCLR pipeline. The original SimCLR expectedly outperforms the other methods because it uses the privileged information injected by the employed data augmentation methods. For example, random cropping is an inductive bias that reflects the spatial regularity of the images. Even though it is possible to strengthen our method and autoencoder-based baselines such as VAEs with such augmentation-based strategies, it still doesn't provide the additional inductive bias of preservation of high-level information in the presence of these augmentations, which SimCLR directly uses. Thus, we restricted all baselines to the generic setting without this inductive bias and leave the domain-specific improvements for future work.

It is seen that the DRL and VDRL methods significantly outperform the baselines on all the datasets at a number of different time-steps $t$. We further evaluate the infinite-dimensional representation on few-shot image classification using the representation at different timescales as input. The detailed results are shown in \Cref{sec:inf_repr_few_shot}. In summary, the representations of DRL and VDRL achieve significant improvements as compared to an autoencoder or VAE for several values of $t$ .

Overall the results align with the theoretical argument of Proposition \ref{theo:infinite_repr} that the rich representation of DRL is at least as good as the static code learned using a reconstruction objective. It further shows that in practice, the infinite-dimensional code is superior to the static (finite-dimensional) representation for downstream applications such as image classification by a significant margin.

As a further analysis, we consider the same experiments when the DRL models are trained on the Variance Preserving SDE formulation \cite{song2021scorebased}. 
\begin{align}
    \diff x = -\frac{1}{2}\beta(t) x \diff t + \sqrt{\beta(t)} \diff \wien,
\end{align}
Figure \ref{fig:vp} shows that even in this formualtion, DRL and VDRL models outperform their autoencoder and denoising autoencoder competitors and perform better than restricted constrastive learning, showing that this approach can be easily adapted to various different diffusion models.
\subsection{Semi-Supervised Image Classification}
\label{sec:downstreamtask}
The current state-of-the-art model for many semi-supervised image classification benchmarks is LaplaceNet \citep{DBLP:journals/corr/abs-2106-04527}. It alternates between assigning pseudo-labels to samples and supervised training of a classifier. The key idea is to assign pseudo-labels by minimizing the graphical Laplacian of the prediction matrix, where similarities of data samples are calculated on a hidden layer representation in the classifier.
Note that LaplaceNet applies \emph{mixup} \citep{DBLP:journals/corr/abs-1710-09412} that changes the input distribution of the classifier. We evaluate our method with and without mixup on CIFAR-10 \citep{cifar10}, CIFAR-100 \citep{cifar100} and MiniImageNet \citep{DBLP:journals/corr/VinyalsBLKW16}.

In the following, we evaluate the infinite-dimensional representation $(E_\phi(x_0, t))_{t\in[0, T]}$ on semi-supervised image classification, where we use DRL and VDRL as pretraining for the LaplaceNet classifier. Table \ref{tab:clf_by_nlabels} depicts the classifier accuracy on test data for different pretraining settings. Details for architecture and hyperparameters are described in \Cref{sec:semisup_architecture}.

Our proposed pretraining using DRL significantly improves the baseline and often surpasses the state-of-the-art performance of LaplaceNet. Most notable are the results of DRL and VDRL without mixup, which achieve high accuracies without being specifically tailored to the downstream task of classification. Note that pretraining the classifier as part of an autoencoder did not yield any improvements (Table \ref{tab:clf_autoenc} in the Appendix).
Combining DRL with mixup yields inconsistent improvements, results are reported in Table \ref{tab:clf_by_nlabels_mixup} of the Appendix. 
In addition, DRL pretraining achieves much better performances when only limited computational resources are available (Tables \ref{tab:clf_by_nlabels_100}, \ref{tab:clf_fewlabels_100} in the Appendix).



\subsection{Qualitative Results}
\label{sec:qual}
We first train a DRL model with $L_1$-regularization on the latent code on MNIST \citep{lecun-mnisthandwrittendigit-2010} and CIFAR-10.
Figure \ref{fig:vis_drl} (\textit{left}) shows samples from a grid over the latent space and a point cloud visualization of the latent values $z = E_\phi(x_0)$. For MNIST, we can see that the value of $z_1$ controls the stroke width, while $z_2$ weakly indicates the class. The latent code of CIFAR-10 samples mostly encodes information about the background color, which is weakly correlated to the class. The use of a probabilistic encoder (VDRL) leads to similar representations, as seen in Fig. \ref{fig:vis_vdrl} (\textit{left}). We further want to point out that the generative process using the reverse SDE involves randomness and thus generates different samples for a single latent representation. The diversity of samples however steadily decreases with the dimensionality of the latent space, shown in Figure \ref{fig:randomness_dimensionality} of the Appendix. 

Next, we analyze the behavior of the representation when adjusting the weighting function $\lambda(t)$ to focus on higher noise levels, which can be done by changing the sampling distribution of $t$.
To this end, we sample $t \in [0, 1]$ such that $\sigma(t)$ is uniformly sampled from the interval $[\sigma_{\min}, \sigma_{\max}] = [0.01, 50]$. 
Figure \ref{fig:vis_drl} (\textit{right}) shows the resulting representation of DRL and Figure \ref{fig:vis_vdrl} (\textit{right}) for the VDRL results. As expected, the latent representation for MNIST encodes information about classes rather than fine-grained features such as stroke width. This validates our hypothesis of Section \ref{sec:control} that we can control the granularity of features encoded in the latent space. For CIFAR-10, the model again only encodes information about the background, which contains the most information about the image class. A detailed analysis of class separation in the extreme case of training on single timescales is included in \Cref{sec:single_t}.

Overall, the difference in the latent codes for varying $\lambda(t)$ shows that we can control the granularity encoded in the representation of DRL. This ability provides a significant advantage when there exists some prior information about the level of detail that we intend to encode in the target representation. We further illustrate how the representation encodes information for the task of denoising in the Appendix (Fig. \ref{fig:denoise_sample}).

We also provide further analysis into the impact of noise scales on generation in Appendix \ref{sec:init_noise_scale}.
\section{Conclusion}
We presented Diffusion-based Representation Learning (DRL), a new objective for representation learning based on conditional denoising score matching. In doing so, we turned the original non-vanishing objective function into one that can be reduced arbitrarily close to zero by the learned representation.
We showed that the proposed method learns interpretable features in the latent space. In contrast to some of the previous approaches that required specialized architectural changes or data manipulations, denoising score matching comes with a natural ability to control the granularity of features encoded in the representation. We demonstrated that the encoder can learn to separate classes when focusing on higher noise levels and encodes fine-grained features such as stroke-width when mainly trained on smaller noise variance.
In addition, we proposed an infinite-dimensional representation and demonstrated its effectiveness for downstream tasks such as few-shot classification. Using the representation learning as pretraining for a classifier, we were able to improve the results of LaplaceNet, a state-of-the-art model on semi-supervised image classification.

Starting from a different origin but conceptually close, contrastive learning as a self-supervised approach could be compared with our representation learning method. We should emphasize that there are fundamental differences both at theoretical and algorithmic levels between contrastive learning and our diffusion-based method. The generation of positive and negative examples in contrastive learning requires the domain knowledge of the applicable invariances. This knowledge might be hard to obtain in scientific domains such as genomics where the knowledge of invariance amounts to the knowledge of the underlying biology which in many cases is not known. However, our diffusion-based representation uses the natural diffusion process that is employed in score-based models as a continuous obfuscation of the information content. Moreover, unlike the loss function of the contrastive-based methods that are specifically designed to learn the invariances of manually augmented data, our method uses the same loss function that is used to learn the score function for generative models. The representation is learned based on a generic information-theoretic concept which is an encoder (information channel) that controls how much information of the input has to be passed to the score function at each step of the diffusion process. We also provided theoretical motivation for this information channel. The algorithm cannot ignore this source of information because it is the only way to reduce a non-negative loss arbitrarily close to zero.

Our experiments on diffusion-based representation learning methods highlight its benefits when compared to \textit{fully} unsupervised models like autoencoders, variational or denoising. The proposed methodology does not rely on additional supervision regarding augmentations, and can be easily adapted to any representation learning paradigm that previously relied on reconstruction-based autoencoder methods.

\section*{Acknowledgements}
SM would like to acknowledge the support of UNIQUE’s and IVADO’s scholarships towards his research. This research was enabled in part by compute resources provided by Mila (mila.quebec).

\clearpage
\bibliography{example_paper}
\bibliographystyle{icml2023}

\newpage
\appendix
\onecolumn
\section{Related work on contrastive learning}
\label[appendix]{sec:contrastive_learning}
The core idea of contrastive learning is to learn representations that are similar for different views of the same image and distant for different images. In order to prevent the collapse of representations to a constant, various approaches have been introduced. SimCLRv2 directly includes a loss term repulsing negative image pairs in addition to the attraction of different views of positive pairs \citep{DBLP:journals/corr/abs-2006-10029}). In contrast, BYOL relies solely on positive pairs, preventing collapse by enforcing similarity between the encoded representation of an image and the output of a momentum encoder applied to a different view of the same image \citep{grill2020bootstrap}. An additional approach relies on online clustering and was proposed in SwAV \citep{caron2021unsupervised}. Training in SwAV is based on enforcing consistency between cluster assignments produced for different views of an image. Each of these methods relies on the foundation of Siamese networks \citep{articlesiamese}, which were shown to be competitive for unsupervised pretraining for classification networks on its own when including a stop-gradient operation on one of the branches \citep{chen2020exploring}.

\section[appendix]{Denoising Score Matching}
The following is the proof for the new formulation of the denoising score matching objective in Equation \ref{eq:dsmesm}.
\begin{proof}
\label{proof:dsm_formulation}
It was shown by \cite{6795935} that Equation \ref{eq:dsm} is equal to explicit score matching up to a constant which is independent of $\theta$, that is,
\begin{align}
    &\mathbf{E}_{x_0} \large\{ \mathbf{E}_{x_t | x_0} \large[
    \lVert s_\theta(x_t, t) - \nabla_{x_t} \log p_{0t}(x_t | x_0) \rVert^2_2 \,\large] \large\}\\
    =\, &\mathbf{E}_{x_t} \left[ \lVert s_\theta(x_t, t) - \nabla_{x_t} \log p_{t}(x_t) \rVert^2_2 \right] + c.
\end{align}

As a consequence, the objective is minimized when the model equals the ground-truth score function $s_\theta(x_t, t) = \nabla_{x} \log p_{t}(x)$.
Hence we have:
\begin{align}
    &\mathbf{E}_{x_0} \large\{ \mathbf{E}_{x_t | x_0} \large[
    \lVert \nabla_{x_t} \log p_{t}(x_t) - \nabla_{x_t} \log p_{0t}(x_t | x_0) \rVert^2_2 \,\large] \large\}\\
    =\, &\mathbf{E}_{x_t} \left[ \lVert \nabla_{x_t} \log p_{t}(x_t) - \nabla_{x_t} \log p_{t}(x_t) \rVert^2_2 \right] + c\\
    =\, &c.
\end{align}
Combining these results leads to the claimed exact formulation of the Denoising Score Matching objective:
\begin{align}
    J^{DSM}_t(\theta) =\, &\mathbf{E}_{x_0} \large\{ \mathbf{E}_{x_t | x_0} \large[
    \lVert s_\theta(x_t, t) - \nabla_{x_t} \log p_{0t}(x_t | x_0) \rVert^2_2 \,\large] \large\}\\
    =\, &\mathbf{E}_{x_t} \left[ \lVert s_\theta(x_t, t) - \nabla_{x_t} \log p_{t}(x_t) \rVert^2_2 \right] + c\\
    \begin{split}
    =\, &\mathbf{E}_{x_t} \left[ \lVert s_\theta(x_t, t) - \nabla_{x_t} \log p_{t}(x_t) \rVert^2_2 \right]\\ &+ \mathbf{E}_{x_0} \large\{ \mathbf{E}_{x_t | x_0} \large[
    \lVert \nabla_{x_t} \log p_{t}(x_t) - \nabla_{x_t} \log p_{0t}(x_t | x_0) \rVert^2_2 \,\large] \large\}
    \end{split}\\
    \begin{split}
    = &\mathbf{E}_{x_0} \{ \mathbf{E}_{x_t|x_0}[
    \begin{aligned}[t]
    &\lVert \nabla_{x_t} \log p_{0t}(x_t | x_0) - \nabla_{x_t} \log p_t(x_t) \rVert^2_2\\ 
    &+ \lVert s_\theta(x_t, t) - \nabla_{x_t} \log p_t(x_t) \rVert^2_2 ] \}.
    \end{aligned}
    \end{split}
\end{align}

\end{proof}

\section{Representation learning}
\label[appendix]{theo:infinite_repr_proof}
Here we present the proof for Proposition \ref{theo:infinite_repr}, stating that the infinite-dimensional code learned using DRL is at least as good as a static code learned using a reconstruction objective.
\begin{proof}
\label{proof:infinite_repr}
We assume that the distribution of the diffused samples at time $t = T$ matches a known prior $p_T(x_T)$. That is, $\int p(x_0) p_{0T}(x_T|x_0) \diff x_0 = p_T(x_T)$. In practice $T$ is chosen such that this assumption approximately holds.\\
Now consider the training objective in Equation \ref{eq:time_repr_obj} at time $T$, which can be transformed to a reconstruction objective in the following way:
\begin{align}
    &\lambda(T)\mathbf{E}_{x_0, x_T}\left[\left\lVert s_\theta(x_T, T, E_\phi(x_0, T)) - \nabla_{x_T}\log p_{0T}(x_T|x_0)\right\rVert^2_2\right]\\
    = &\lambda(T) \mathbf{E}_{x_0} \mathbf{E}_{x_T \sim p_T(x_T)}\left[\left\lVert s_\theta(x_T, T, E_\phi(x_0, T)) - \frac{x_0 - x_T}{\sigma^2(T)}\right\rVert^2_2\right]\\
    = &\lambda(T) \sigma^{-4}(T)\mathbf{E}_{x_0} \mathbf{E}_{x_T \sim p_T(x_T)}\left[\left\lVert D_\theta(E_\phi(x_0, T)) - x_0\right\rVert^2_2\right]\\
    = &\lambda(T) \sigma^{-4}(T)\mathbf{E}_{x_0}\left[\left\lVert D_\theta(E_\phi(x_0, T)) - x_0\right\rVert^2_2\right],
\end{align}
where we replaced the score model with
a Decoder model $s_\theta(x_T, T, E_\phi(x_0, T)) = \frac{D_\theta(E_\phi(x_0, T)) - x_T}{\sigma^2(T)}$ and replaced the score function of the perturbation kernel $\nabla_{x_T}\log p_{0T}(x_T|x_0)$ with its known closed-form solution $\frac{x_0 - x_T}{\sigma^2(T)}$ determined by the Forward SDE in Equation \ref{eq:forward_sde}.
Hence the learned code at time $t=T$ is equal to a code learned using a reconstruction objective.

We model a downstream task as a minimization problem of a distance $d: \Omega \times \Omega \to \mathbb{R}$ in the feature space $\Omega$ between the true feature extractor $g: \mathbb{R}^d \to \Omega$ which maps data samples $x_0$ to a features space $\Omega$ and a model feature extractor $h_\psi: \mathbb{R}^c \to \Omega$ doing the same given the code as input.
The following shows that the infinite-dimensional representation is at least as good as the static code:
\begin{equation}
    \inf_t \min_\psi \mathbf{E}_{x_0} [d(h_\psi(E_\phi(x_0, t)), g(x_0))] \leq \min_\psi \mathbf{E}_{x_0} [d(h_\psi(E_\phi(x_0, T)), g(x_0))]
\end{equation}
\end{proof}

\begin{figure}[ht]
\vskip 0.2in
\centering
\includegraphics[width=0.5\columnwidth]{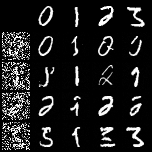}
\caption{Samples generated starting from $x_t$ (left column) using the diffusion model with the latent code of another $x_0$ (top row) as input. It shows that samples are denoised correctly only when conditioning on the latent code of the corresponding original image $x_0$.}
\label{fig:denoise_sample}
\end{figure}

\begin{figure}[ht]
\vskip 0.2in
\centering
\subfigure[$2$-dimensional]{\includegraphics[width=0.35\columnwidth]{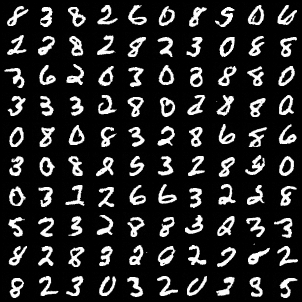}}
\hspace{2mm}
\subfigure[$4$-dimensional]{\includegraphics[width=0.35\columnwidth]{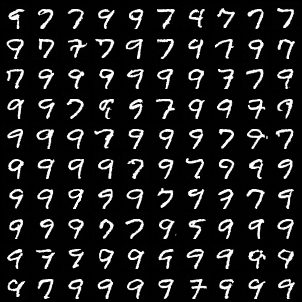}}
\vskip\baselineskip
\subfigure[$8$-dimensional]{\includegraphics[width=0.35\columnwidth]{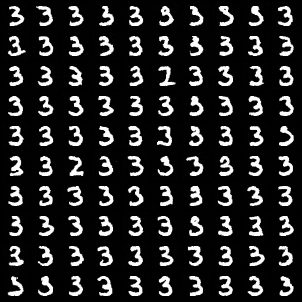}}
\hspace{2mm}
\subfigure[$16$-dimensional]{\includegraphics[width=0.35\columnwidth]{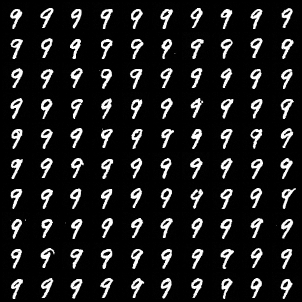}}
\caption{Samples generated using the same latent code for each generation, showing that the randomness of the code-conditional generation of DRL reduces in higher dimensional latent spaces.}
\label{fig:randomness_dimensionality}
\end{figure}

\section{Architecture and Hyperparameters}
\label[appendix]{hyperparams}
The model architecture we use for all experiments is based on “DDPM++ cont. (deep)” used for CIFAR-10 in \cite{song2021scorebased}. It is composed of a downsampling and an upsampling block with residual blocks at multiple resolutions. We did not change any of the hyperparameters of the optimizer. Depending on the dataset, we adjusted the number of resolutions, number of channels per resolution, and the number of residual blocks per resolution in order to reduce training time.

For representation learning, we use an encoder with the same architecture as the downsampling block of the model, followed by another three dense layers mapping to a low dimensional latent space. Another four dense layers map the latent code back to a higher-dimensional representation. It is then given as input to the model in the same way as the time embedding. That is, each channel is provided with a conditional bias determined by the representation and time embedding at multiple stages of the downsampling and upsampling block.

\paragraph{Regularization of the latent space}
For both datasets, we use a regularization weight of $10^{-5}$ when applying L1-regularization, and a weight of $10^{-7}$ when using a probabilistic encoder regularized with KL-Divergence.

\paragraph{MNIST hyperparameters}
\label{mnist_hyperparams}
Due to the simplicity of MNIST, we only use two resolutions of size $28\times28\times32$ and $14\times14\times64$, respectively. The number of residual blocks at each resolution is set to two. In each experiment, the model is trained for $80k$ iterations. For a uniform sampling of $\sigma$ we trained the models for an additional $80k$ iterations with a frozen encoder and uniform sampling of $t$.

\paragraph{CIFAR-10 hyperparameters} 
\label{cifar_hyperparams_small}
For the silhouette score analysis, we use three resolutions of size $32\times32\times32$, $16\times16\times32$, and $8\times8\times32$, again with only two residual blocks at each resolution. Each model is trained for $90k$ iterations.

\paragraph{CIFAR-10 (deep) hyperparameters} 
\label{cifar_hyperparams}
While representation learning works for small models already, sample quality on CIFAR-10 is poor for models of the size described above. Thus for models used to generate samples, we use eight residual blocks per resolution and the following resolutions: $32\times32\times32$, $16\times16\times64$, $8\times8\times64$, and $4\times4\times64$. Each model is trained for $300k$ iterations. Note that this number of iterations is not sufficient for convergence, however capable of illustrating the representation learning with limited computational resources.

\section{Evaluation of the infinite-dimensional representation}
\label[appendix]{sec:inf_repr_few_shot}

\begin{figure}[t!]
\centering
\subfigure[100 labels]{\includegraphics[width=0.45\columnwidth]{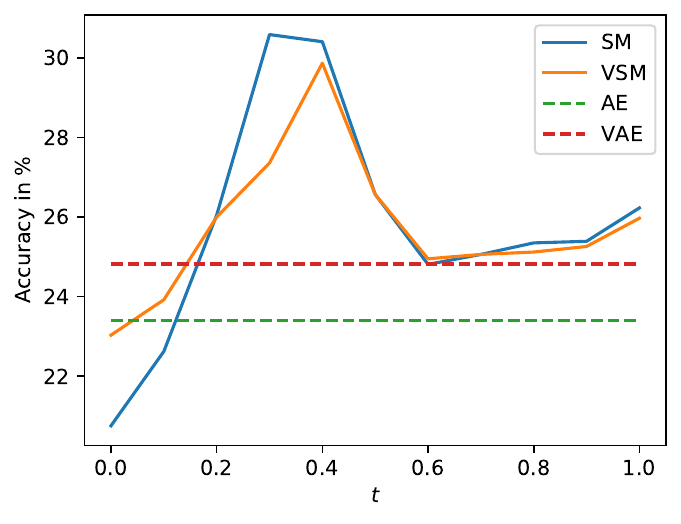}}
\hspace{0.2mm}
\subfigure[1000 labels]{\includegraphics[width=0.45\columnwidth]{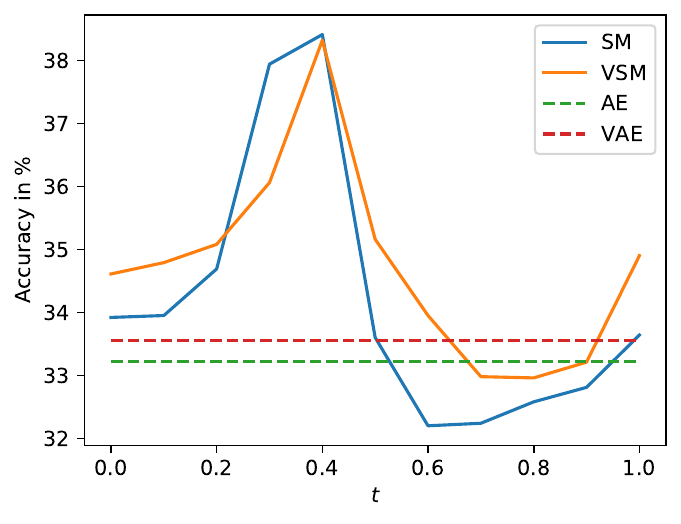}}
\caption{Classifier accuracies for few shot learning on given $8$-dimensional representations learned using DRL (SM), VDRL (VSM), Autoencoder (AE) and Variational Autoencoder (VAE).}
\label{fig:repr_few_shot}
\end{figure}

In order to evaluate our infinite-dimensional representation, we conduct an ablation study where we compare our proposed method with Autoencoders (AE) and Variational Autoencoders (VAE) on CIFAR-10 images. We measure the accuracy of an SVM provided by sklearn \citep{scikit-learn} with default hyperparameters trained on the representation of 100 (resp. 1000) training samples and their class labels. For our time-dependent representation, this is done for fixed values of $t$ between $0.0$ and $1.0$ in steps of $0.1$. This is done for both DRL and VDRL, where we use a probabilistic encoder regularized by including an additional KL-Divergence term in the training objective. DRL and AE were regularized using L1-norm, and the regularization weight was optimized for each model independently.

Results for few-shot learning with fixed representations are shown in Figure \ref{fig:repr_few_shot}. As expected, the accuracies when training on the score matching representations highly depend on the value of $t$. Overall our representation achieves much better scores when using the best $t$, and performs comparable to AE and VAE for $t=1.0$. This aligns with Proposition \ref{theo:infinite_repr} claiming that our representation learning method for $t=1.0$ is similar to a static code learned using reconstruction objective.
Note that the shape of the time-dependent classifier accuracies resembles the one of the silhouette scores of CIFAR-10 in \ref{fig:silscores}. This is not surprising, since both training on single values of $t$ and learning a time-dependent representation are both trained to find the optimal representation for a given value of $t$. We further want to point out that representation learning through score matching enjoys the training stability of diffusion-based generative models, which is often not the case in GANs and VAEs.

\begin{figure*}[t!]
\centering
\includegraphics[width=\textwidth]{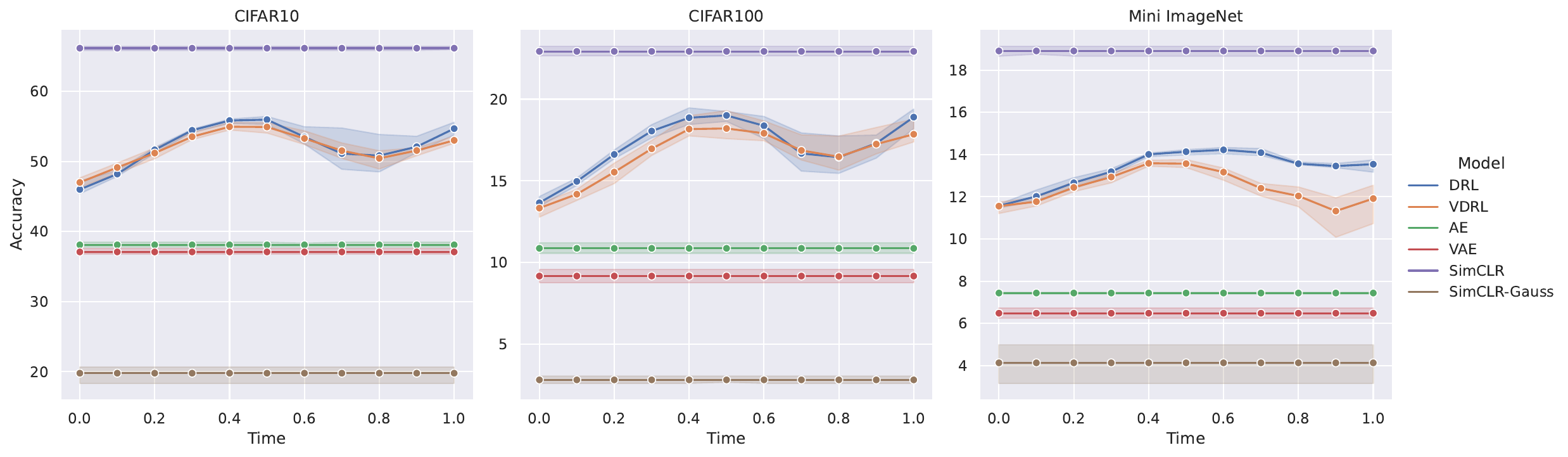}
\caption{Comparing the low-data regime (1000 labels) downstream performance of the proposed diffusion-based representations (DRL and VDRL) with the baselines that include autoencoder (AE), variational autoencoder (VAE), simple contrastive learning (simCLR) and its restricted variant (simCLR-Gauss) which exclude domain-specific data augmentation from the original simCLR algorithm.}
\label{fig:perf_1000}
\end{figure*}

\begin{figure*}[t!]
\centering
\includegraphics[width=\textwidth]{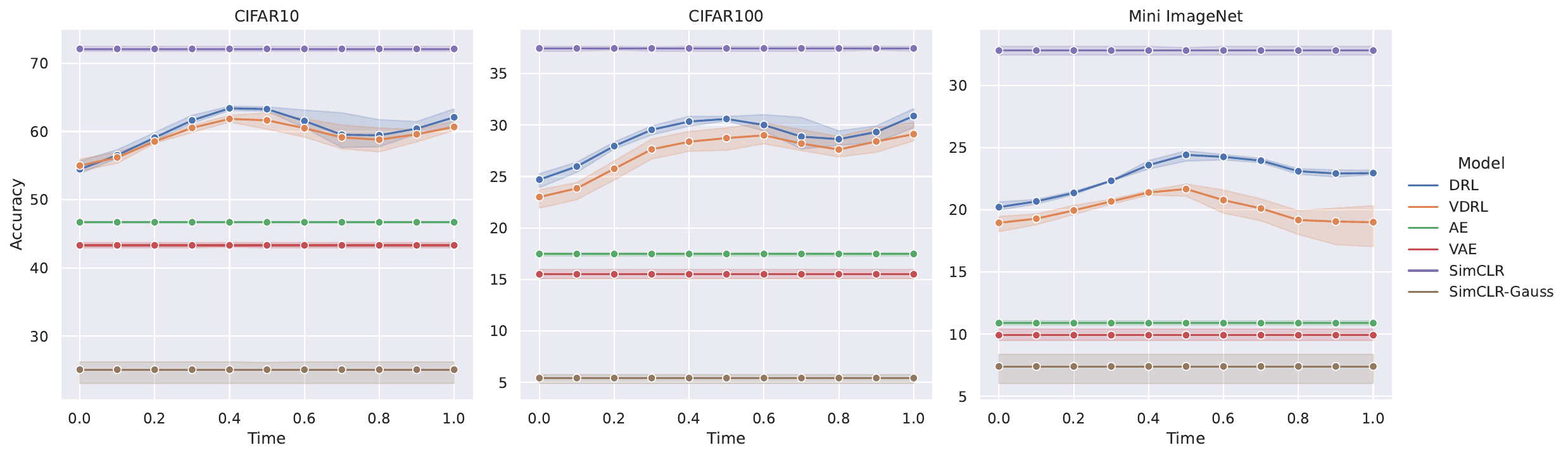}
\caption{Comparing the low-data regime (5000 labels) downstream performance of the proposed diffusion-based representations (DRL and VDRL) with the baselines that include autoencoder (AE), variational autoencoder (VAE), simple contrastive learning (simCLR) and its restricted variant (simCLR-Gauss) which exclude domain-specific data augmentation from the original simCLR algorithm.}
\label{fig:perf_5000}
\end{figure*}

\begin{figure*}[t!]
\centering
\includegraphics[width=\textwidth]{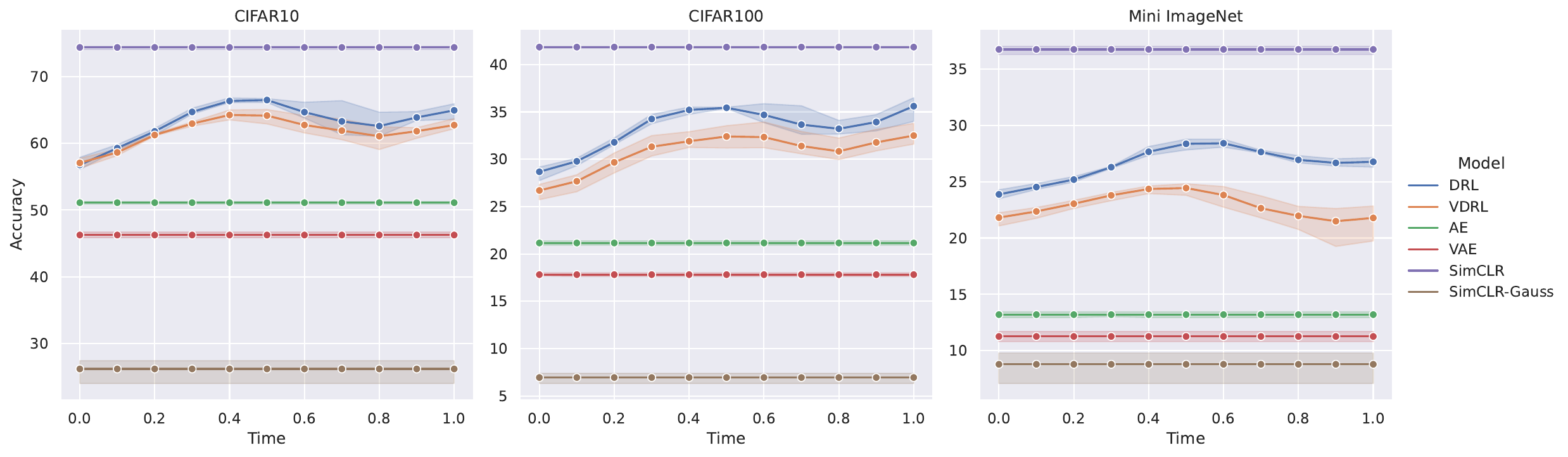}
\caption{Comparing the low-data regime (10000 labels) downstream performance of the proposed diffusion-based representations (DRL and VDRL) with the baselines that include autoencoder (AE), variational autoencoder (VAE), simple contrastive learning (simCLR) and its restricted variant (simCLR-Gauss) which exclude domain-specific data augmentation from the original simCLR algorithm.}
\label{fig:perf_10000}
\end{figure*}

\section{Downstream Image Classification}
\label[appendix]{sec:downstream_arch}
\paragraph{Architecture and Hyperparameters} In all our experiments, we consider the small WideResNet model WRN-28-2 of \cite{DBLP:journals/corr/abs-2106-04527} as the encoder module for all of the different settings: diffusion representation learning, autoencoder and contrastive learning. We sample the time-steps at intervals of $0.1$ from the range $0.0 - 1.0$. Corresponding to each time-step, we train a single layered non-linear MLP network for 50 epochs.

\paragraph{Results with Limited Data} We perform additional experiments where the encoder system is as before and kept frozen, but the MLP can only access a fraction of the training set for the downstream supervised classification task. We ablate over three different number of labels provided to the MLP: 1000, 5000 and 10000. The results for the different datasets can be seen in Figures \ref{fig:perf_1000}-\ref{fig:perf_10000} which shows that the trends are consistent even in low data regime.

\section{Semi-supervised image classification}
\label[appendix]{sec:semisup_architecture}

\paragraph{Architecture and Hyperparameters}  In all experiments, our encoder has the same architecture as the classifier, where the hidden layer used to measure similarities for assigning pseudo-labels in LaplaceNet is used as the latent code in representation learning. For all experiments, the input $t$ to the encoder is included as a trainable parameter of the model and initialized with $t = 0.5$. As done in the original paper, we train the model for 260 iterations, where each iteration consists of assigning pseudo-labels and one epoch of supervised training on the assigned pseudo-labels. The training is preceded by 100 supervised epochs on the labeled data. We use the small WideResNet model WRN-28-2 of \cite{DBLP:journals/corr/abs-2106-04527} and the same hyperparameters as the authors. 

\paragraph{Evaluation with limited computation time} In the following we include more detailed analysis of the scenario of a few supervised labels and limited computational resources. Besides LaplaceNet and its version without mixup, we include an ablation study of encoder pretraining as part of an autoencoder using binary cross-entropy as a reconstruction objective. In addition, we propose to improve the search for the optimal value of $t$ by the model selection, since the gradient for $t$ is usually noisy and small. Thus we include additional experiments where we chose the initial $t$ based on the minimum training loss after 100 epochs of supervised training. The optimal $t$ is approximated by calculating the training loss for 11 equally spaced values of $t$ in the interval $[0.001, 1]$. The results are shown in Table \ref{tab:clf_fewlabels_100}. While mixup achieves no significant improvement in the few-label case trained using 100 epochs, we can see that a simple autoencoder pretraining consistently improves classifier accuracy. More notably, however, our proposed pretraining based on score matching achieves significantly better results than both random initialization and autoencoder pretraining. In the $t$-search, we observed that for all datasets, our proposed method selects $t = 0.9$, however it moves towards the interval $[0.4, 0.6]$ during training. While this shows that the approach of selecting $t$ based on supervised training loss is not working, it demonstrates that the parameter $t$ can very well be learned in the training process, making the downstream task performance robust to the initial value of $t$. In our experiments the final value of $t$ was always in the range $[0.4, 0.6]$, independent of the initial value of $t$.

\begin{table*}\centering
\begin{tabular}{@{}lrccccccc@{}}\toprule
Dataset & \#labels & \phantom{ab} & 
No pretraining & \phantom{ab} & 
\begin{tabular}{@{}c@{}}Pretraining using \\ DRL\end{tabular} & \phantom{ab} & 
Improvement
\\
\midrule
CIFAR-10 & 100 && 64.12 && 69.79 && +5.67\\
& 500 && 86.24 && 88.28 && +2.04\\
& 1000 && 87.48 && 88.56 && +1.08\\
& 2000 && 89.99 && 89.52 && -0.47\\
& 4000 && 90.15 && 91.13 && +0.98\\
\midrule
CIFAR-100 & 1000 && 45.14 && 48.04 && +2.90\\
& 4000 && 59.86 && 60.34 && +0.48\\
& 10000 && 64.83 && 65.80 && +0.97\\
& 20000 && 65.77 && 66.39 && +0.62\\
\midrule
MiniImageNet & 4000 && 47.18 && 50.75 && +3.57\\
& 10000 && 58.66 && 58.62 && -0.04\\
\bottomrule
\end{tabular}
\caption{Classifier accuracy in \% with and without DRL as pretraining of the classifier when training for 100 epochs only.}
\label{tab:clf_by_nlabels_100}
\end{table*}

\begin{table*}\centering
\begin{tabular}{@{}llcccccc@{}}\toprule
Pretraining & Options & \phantom{abc} &
\begin{tabular}{@{}c@{}}CIFAR-10 \\ 100 labels\end{tabular} & \phantom{abc} &
\begin{tabular}{@{}c@{}}CIFAR-100 \\ 1000 labels\end{tabular} & \phantom{abc} &
\begin{tabular}{@{}c@{}}MiniImageNet \\ 4000 labels\end{tabular}\\
\midrule
None & && 64.12 && 45.14 && 47.18\\
None & mixup && 54.06 && 46.28 && 47.64\\
DRL & && \textbf{69.79} && \textbf{48.04} && \textbf{50.75}\\
DRL & $t$-search && 67.07 && 47.08 && 50.31\\
Autoencoder & && 64.99 && 46.88 && 48.52\\
\bottomrule
\end{tabular}
\caption{Comparison of classifier accuracy in \% for different pretraining methods in the case of few supervised labels when training for 100 epochs only.}
\label{tab:clf_fewlabels_100}
\end{table*}

\begin{table*}\centering
\begin{tabular}{@{}llcccccc@{}}\toprule
Pretraining & \phantom{abc} &
\begin{tabular}{@{}c@{}}CIFAR-10 \\ 100 labels\end{tabular} & \phantom{abc} &
\begin{tabular}{@{}c@{}}CIFAR-100 \\ 1000 labels\end{tabular} & \phantom{abc} &
\begin{tabular}{@{}c@{}}MiniImageNet \\ 4000 labels\end{tabular}\\
\midrule
None && 73.68 && 55.58 && 58.40\\
DRL && \textbf{74.31} && \textbf{55.85} && \textbf{58.95}\\
Autoencoder && 58.84 && 55.41 && 57.93\\
\bottomrule
\end{tabular}
\caption{Classifier accuracy in \% for autoencoder pretraining compared with the baseline and score matching as pretraining. No mixup is applied for this ablation study.}
\label{tab:clf_autoenc}
\end{table*}

\begin{table*}\centering
\begin{tabular}{@{}lrccccccccccc@{}}\toprule
&&& Ours && Ours && Ours && Ours && Ours 
\\
\multicolumn{2}{@{}l}{Pretraining} && 
Basic && Basic && Mixup-DRL && VDRL && VDRL 
\\
\multicolumn{2}{@{}l}{Mixup in sup. training} && No && Yes && Yes && No && Yes
\\
\midrule
Dataset & \#labels & 
\\
\midrule
CIFAR-10 & 100 && 74.31 && 64.67 && 70.40 && \textbf{81.63} && 77.51\\
& 500 && 92.70 && 92.31 && 92.55 && \textbf{92.79} && 91.46\\
& 1000 && 93.24 && 93.42 && 93.14 && \textbf{93.60} && 93.33\\
& 2000 && \textbf{94.18} && 93.91 && 93.80 && 93.96 && 94.27\\
& 4000 && 94.75 && \textbf{95.22} && 94.75 && 95.00 && 94.87\\
\midrule
CIFAR-100 & 1000 && 55.85 && 55.74 && 55.15 && \textbf{56.47} && 55.65\\
& 4000 && 67.22 && 67.47 && 67.09 && \textbf{67.54} && 67.52\\
& 10000 && 73.31 && 73.66 && \textbf{74.36} && 73.50 && 73.20\\
& 20000 && 76.46 && 76.88 && \textbf{77.04} && 76.64 && 76.68\\
\midrule
MiniImageNet & 4000 && 58.95 && 59.29 && \textbf{59.46} && 59.14 && 59.36\\
& 10000 && 67.31 && 66.63 && 67.31 && \textbf{67.46} && 66.79\\
\bottomrule
\end{tabular}
\caption{Evaluation of classifier accuracy in \%, including the setting of using mixup during pretraining (right column). DRL pretraining is our proposed representation learning, and "Mixup-DRL" the respective version which additionally applies mixup during pretraining. "VDRL" instead uses a probabilistic encoder.}
\label{tab:clf_by_nlabels_mixup}
\end{table*}
\section{Training on single timescales}
\label[appendix]{sec:single_t}
To understand the effect of training DRL on different timescales more clearly, we limit the support of the weighting function $\lambda(t)$ to a single value of $t$. We analyze the resulting quality of the latent representation for different values of $t$ using the silhouette score with euclidean distance based on the dataset classes \cite{ROUSSEEUW198753}. It compares the average distance between a point to all other points in its cluster with the average distance to points in the nearest different cluster. Thus we measure how well the latent representation encodes classes, ignoring any other features. Note that after learning the representation with a different distribution of $t$ it is necessary to perform additional training with a uniform sampling of $t$ and a frozen encoder to achieve good sample quality.

\begin{figure}[t!]
\centering
\subfigure[MNIST]{\label{fig:silscore_mnist}\includegraphics[width=0.45\columnwidth]{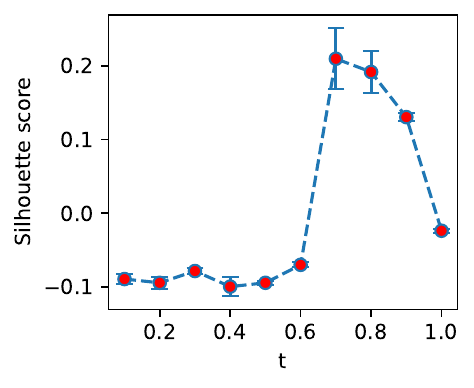}}
\hspace{0.2mm}
\subfigure[CIFAR-10]{\label{fig:silscore_c3}\includegraphics[width=0.45\columnwidth]{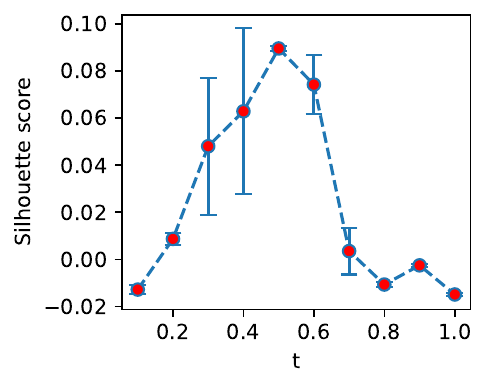}}
\caption{Mean and standard deviation of silhouette scores when training a DRL model on MNIST (left) and CIFAR-10 (right) using a single $t$ over three runs.}
\label{fig:silscores}
\end{figure}

Figure \ref{fig:silscores} shows the silhouette scores of latent codes of MNIST and CIFAR-10 samples for different values of $t$. In alignment with our hypothesis of Section \ref{sec:control}, training DRL on a small $t$ and thus low noise levels leads to almost no encoded class information in the latent representation, while the opposite is the case for a range of $t$ which differs between the two datasets. The decline in encoded class information for high values of $t$ can be explained by the vanishing difference between distributions of perturbed samples when $t$ gets large. This shows that the distinction among the code classes represented by the silhouette score is controlled by $\lambda(t)$.

\begin{table}[t!]
\centering
\begin{tabular}{@{}cccccc@{}}\toprule
$t_{\rm init}$ & $\sigma(t_{\rm init})$ & 
\phantom{ab} & 
\begin{tabular}{@{}c@{}}Gaussian \\ FID $\downarrow$ \end{tabular} & 
\phantom{ab} & 
\begin{tabular}{@{}c@{}}Uniform + Gaussian \\ FID $\downarrow$ \end{tabular}
\\
\midrule
0.5 & 0.71 && 218.95 && 25.02\\
0.6 & 1.66 && 75.11 && 5.15\\
0.7 & 3.88 && 12.57 && 2.98\\
0.8 & 9.10 && 3.05 && 2.99\\
0.9 & 21.33 && 2.97 && 2.94\\
1.0 & 50.00 && 3.01 && 2.99\\
\bottomrule
\end{tabular}
\caption{FID for different initial noise scales evaluated on 20k generated samples.}
\label{tab:init_noise_scale}
\vspace{-4mm}
\end{table}
\begin{figure}[t!]
\centering
\subfigure[\scriptsize $t_{\rm init}=0.5$]{\includegraphics[width=0.15\columnwidth]{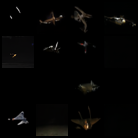}}
\subfigure[$t_{\rm init}=0.6$]{\includegraphics[width=0.15\columnwidth]{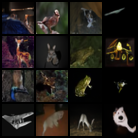}}
\subfigure[$t_{\rm init}=0.7$]{\includegraphics[width=0.15\columnwidth]{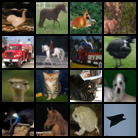}}
\subfigure[$t_{\rm init}=0.8$]{\includegraphics[width=0.15\columnwidth]{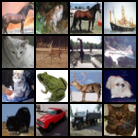}}
\subfigure[$t_{\rm init}=0.9$]{\includegraphics[width=0.15\columnwidth]{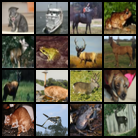}}
\subfigure[$t_{\rm init}=1.0$]{\includegraphics[width=0.15\columnwidth]{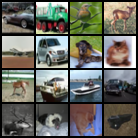}}
\vspace{0.2mm}
\subfigure[$t_{\rm init}=0.5$]{\includegraphics[width=0.15\columnwidth]{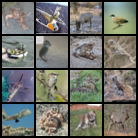}}
\subfigure[$t_{\rm init}=0.6$]{\includegraphics[width=0.15\columnwidth]{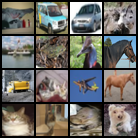}}
\subfigure[$t_{\rm init}=0.7$]{\includegraphics[width=0.15\columnwidth]{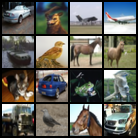}}
\subfigure[$t_{\rm init}=0.8$]{\includegraphics[width=0.15\columnwidth]{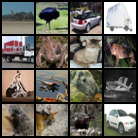}}
\subfigure[$t_{\rm init}=0.9$]{\includegraphics[width=0.15\columnwidth]{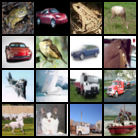}}
\subfigure[$t_{\rm init}=1.0$]{\includegraphics[width=0.15\columnwidth]{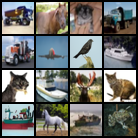}}
\caption{Generated image samples for different values of $t_{\rm init}$. Top row ((a)-(f)) uses the Gaussian prior, bottom row ((g)-(l)) uses the version with an additional uniform random variable in the prior.}
\label{fig:appendix_t_init_samples}
\vskip -0.2in
\end{figure}
\section{The choice of the initial noise scale}
\label{sec:init_noise_scale}
In the following, we evaluate image quality and diversity for different initial noise scales for CIFAR-10 dataset. Note that we do not change $\sigma(T)$, but instead evaluate generated images for different initial times $t_{\rm init}$, which implicitly define the initial noise scale $\sigma(t_{\rm init})$. This reduces the number of sampling steps per image, which is $1000 \times t_{\rm init}$ and thus directly proportional to $t_{\rm init}$.  
Table \ref{tab:init_noise_scale} shows the FID of generated images for various values of $t_{\rm init}$. As we can see, the first 200 sampling steps can safely be replaced by approximating the prior directly either with the Gaussian or the additional uniform distribution. Interestingly, using the sum of the uniform and Gaussian random variables as a prior leads to improved image quality. This approximation for $p_{0.7}(x)$ allows us to reduce the number of sampling steps by 30\% without sacrificing image quality, which is further supported by the visual quality of generated samples shown in Figure \ref{fig:appendix_t_init_samples}. Further, note that FID is occasionally lower for values of $t_{\rm init} < 1.0$ than for $t_{\rm init} = 1$. This suggests that up to these timescales, our prior approximates the distribution better than the diffusion model when starting at $t_{\rm init} = 1.0$. 
\end{document}